\documentclass[11pt,a4paper]{article}
\usepackage[margin=2.5cm]{geometry}
\usepackage{times}
\usepackage{graphicx}
\usepackage{booktabs}
\usepackage{longtable}
\usepackage{array}
\usepackage{amsmath}
\usepackage{hyperref}
\usepackage{caption}
\usepackage{titlesec}
\usepackage{enumitem}
\usepackage{fancyhdr}
\usepackage[table]{xcolor}
\usepackage[numbers,sort&compress]{natbib}
\usepackage{ragged2e}
\usepackage{float}
\usepackage{pifont}   
\usepackage{pdfpages}

\hypersetup{colorlinks=true, linkcolor=blue, citecolor=blue, urlcolor=blue}

\titleformat{\section}{\large\bfseries}{\thesection}{1em}{}
\titleformat{\subsection}{\normalsize\bfseries}{\thesubsection}{1em}{}

\newcolumntype{L}[1]{>{\raggedright\arraybackslash}p{#1}}

\title{\textbf{\justifying Machine Learning-Based Prediction of Dengue Infection Using Clinical and Hematological Data from Bangladesh: A Comparative Analysis of Benchmark and Extended Prediction Models}}
\author{
  Neesha Naimeen \\ 
  School of Engineering, Computer and Mathematical Sciences\\ 
  Auckland University of Technology (AUT)\\
  Auckland,New Zealand\\
  ztn6203@autuni.ac.nz\\
  \and
  Victor Miranda Soberanis \\ 
  Department of Mathematical Sciences\\ Auckland University of Technology (AUT)\\ 
  Auckland,New Zealand\\
  victor.miranda@aut.ac.nz 
}
\date{}

\begin{document}

\maketitle

\begin{abstract}
\noindent Dengue is a viral infection caused by the dengue virus (DENV) and a major public health
concern in Bangladesh, where symptom overlap with other febrile illnesses complicates
early diagnosis. This study compares Logistic Regression, Random Forest and XGBoost
to predict dengue from clinical and hematological data using two Bangladeshi hospital
datasets (2024, \textit{n}=1{,}511; 2025, \textit{n}=1{,}283) with SMOTE. Extended models outperformed
benchmarks within 2024 (Random Forest ROC-AUC 0.640 to 0.661), with RDW-CV the
most consistently influential added predictor. 2025 benchmark performance was higher
(ROC-AUC 0.904), unexplained by class imbalance and possibly reflecting data quality.
External validation is needed.

\vspace{0.5em}
\end{abstract}

\section{Introduction}
Dengue fever is a major mosquito-borne disease and a growing public health concern in Bangladesh.The
country experienced its largest and deadliest outbreak on record in 2023, when the Directorate General
of Health Services (DGHS),Ministry of Health and Family Welfare, Government of Bangladesh, reported 321,179 hospitalised cases and 1,705 deaths nationwide - 1.3 times more
cases and twice as many deaths as recorded across the previous 23 years combined \citep{hasan2025dengue}. Independently,
the World Health Organization confirmed 69,483 laboratory-confirmed cases and 327 deaths nationally between 1 January and 7 August 2023 alone(a case-fatality rate of 0.47\%), with 63\% of all cases and 62\%
of all deaths for the period reported within July 2023 alone as the outbreak accelerated sharply \citep{who2023dengue}. Rising temperatures, increased rainfall, rapid urbanisation and other climate-related factors have contributed to more frequent and severe dengue outbreaks by creating favourable conditions for Aedes mosquitoes. Early detection of dengue is essential to reduce disease severity, improve patient outcomes and support
timely clinical intervention.In a 2022 Bangladeshi cohort, 4.2\% of hospitalised dengue patients died, with hypotension, rising haematocrit and falling platelet count during the febrile phase significantly predicting progression to severe dengue \citep{sami2023dengue}, underscoring the value of identifying such warning signs as early as possible.

Machine learning (ML) has increasingly been used to help diagnose diseases and several recent studies have used it for dengue prediction with clinical and blood-test data from Bangladeshi hospitals. \citet{assaduzzaman2024} and \citet{nirob2025} published datasets of hematological data for dengue research in Bangladesh and \citet{tuhin2025} used same \cite{nirob2025} data to build a stacking ensemble model that reported high accuracy. \citet{haque2025} combined genetic algorithm-based feature selection with GAN-generated synthetic data to deal with class imbalance, while \citet{hossain2025} described the clinical and blood-test patterns seen in confirmed dengue cases, without building a prediction model. Similar work has also been done outside Bangladesh~\cite{ansari2024,dasgupta2025, marandi2023, qaiser2024,bohm2024} showing that blood-test results are useful for predicting dengue in different countries too. However, few of these studies have systematically examined how adding more clinical and blood-test features affects the accuracy of ML models and fewer still evaluate how well such models generalise beyond the single dataset on which they were built.

Diagnosing dengue early is also difficult because its symptoms look like other common fevers which can delay treatment and put more pressure on hospitals. Even though hospitals routinely collect clinical and blood-test data, relatively little research in Bangladesh \cite{assaduzzaman2024,nirob2025,tuhin2025, haque2025,hossain2025} has looked at whether adding more of these lab measurements actually improves how well machine learning models predict dengue and whether these improvements are consistent across different patient groups and time periods. This study addresses that gap through three contributions: (i) comparing machine learning models trained on the
predictors common to two independently collected Bangladeshi datasets; (ii) assessing the added value of the further hematological predictors available in the 2024 dataset and (iii) evaluating how consistently model performance generalises across both datasets which were collected at different hospitals in
different years.

\section{Literature Review}
\subsection{Studies Using Bangladeshi Clinical and Hematological Datasets}
\label{subsec:bangladeshi_studies}

Several recent studies have applied machine learning to dengue diagnosis using clinical and hematological data, with a particular concentration of work using Bangladeshi hospital datasets closely related to those used in this study.

\citet{assaduzzaman2024} published a benchmark dataset of 1,003  dengue patient records   from Upazila Health Complex, Kalai, Jaipurhat, Bangladesh, recording demographic and hematological parameters including age, sex, hemoglobin, WBC count, differential white cell counts, the RBC panel, platelet count and PDW. Their statistical analysis found a significant gender-based difference in hemoglobin levels, illustrating that demographic factors interact with hematological presentation in dengue and motivating the inclusion of Age and Sex as predictors in the present study. 

\citet{nirob2025} extended this line of work with a more comprehensive hematological dataset for dengue incidence, sourced from Jamalpur Medical College Hospital, Jamalpur, Bangladesh, comprising 1,523 patient records. Their statistical testing found several hematological parameters including Age ($p = 0.0004$), Neutrophils\%, Lymphocytes\%, Monocytes\%, Eosinophils\%, and RBC count (all $p<0.001$) to differ significantly between dengue-positive and dengue-negative patients. This supports the idea that routine blood test results can help diagnose dengue on their own without needing specialized serological tests.

\citet{tuhin2025} built directly on this same 1,523-patient dataset to develop \textit{DengueStackX-19}, a stacking ensemble model combined with SMOTEENN for class-imbalance correction. Their model achieved 93.65\% accuracy and 89.63\% F1-score under 10-fold cross-validation, rising to 96.38\% accuracy in a further evaluation phase. They additionally applied SHAP and LIME to identify the hematological features most influential to model predictions, reporting that the resulting importance patterns aligned with established clinical understanding of dengue pathophysiology and deployed the model as a web-based diagnostic tool. This study is the most directly comparable prior work to the present one, since it uses the same source data as the 2024 dataset here; the substantially higher accuracy they report (93--96\%) relative to the results obtained in this study (Section~\ref{sec:results}) is a useful point of comparison discussed further in Section~\ref{sec:discussion} and may partly reflect differences in feature engineering, ensembling strategy and the specific imbalance-correction method used (SMOTEENN, which combines oversampling with edited nearest-neighbour cleaning, versus the plain SMOTE used here).

\citet{haque2025} proposed an alternative approach combining Genetic Algorithm-based feature selection with GAN-based synthetic data generation to address class imbalance, paired with explainable AI (XAI) methods for interpretability. Their optimized Decision Tree model achieved 99.49\% testing accuracy and the framework was further optimized for near-real-time inference with an operational response latency under 0.6 seconds, deployed via a web interface intended for outbreak-management settings. While such a high accuracy figure is notable, it also raises the possibility of overfitting to a specific localized dataset, a caution relevant to interpreting very high accuracy claims in this literature more broadly.

\citet{hossain2025} took a more clinically descriptive approach, conducting a cross-sectional study of dengue patients treated during Bangladesh's 2022 outbreak at a tertiary care hospital, characterising the clinical and hematological manifestation of confirmed cases rather than building a predictive model. This work is useful as clinical grounding for interpreting which hematological changes are expected during dengue infection, complementing the more model-focused studies above.

\subsection{International Applications of Machine Learning to Dengue Prediction}
\label{subsec:international_studies}

In India, \citet{ansari2024} used machine learning to predict blood component transfusion needs in hospitalized dengue patients across a private hospital group, while \citet{dasgupta2025} derived simple mathematical equations from five elementary laboratory parameters- age, total leucocyte count, hemoglobin, platelet count and erythrocyte sedimentation rate  to classify dengue positivity, reporting that platelet count was the most influential predictor, a finding consistent with the present study. 

In Sri Lanka, \citet{marandi2023} developed a random forest and LightGBM ensemble to predict plasma leakage from early-infection laboratory parameters, including lymphocyte count, hemoglobin, hematocrit, age and aspartate aminotransferase, evaluated on a genuinely held-out test set rather than cross-validation alone.

In Pakistan, \citet{qaiser2024} compared several machine learning models on routine clinical data and found that a support vector machine outperformed other approaches for early dengue diagnosis. 

In Brazil, \citet{bohm2024} applied machine learning to dengue case screening, identifying which routinely collected clinical variables were most useful for triage decisions. 

\subsection{Research Gaps}
\label{subsec:research_gaps}

Overall, this international work shows that blood test results can help predict dengue fairly well in different countries, not just Bangladesh. But like the Bangladeshi studies, most of these papers only look at one group of patients from one place and time. None of them test how a model trained on one group performs on a genuinely different population or time period later on.

Looking at all these studies together, three gaps stand out for this study to address. First, most studies use only one dataset from one hospital and one time period. This means we don't know how well a model would work on a different group of patients or a different time period. This is exactly why this study compares 2024 and 2025 data directly. Second, some studies deal with imbalanced classes using methods like SMOTEENN or GAN-based data generation, but few of them compare oversampling and undersampling side by side to show the trade-off between the two, this study does that directly (Section~\ref{sec:results}). Third, no study found so far compares a "benchmark" set of predictors against an "extended" set of predictors on the same group of patients to see exactly how much the extra predictors help.This is the main comparison this study makes (Section~\ref{subsec:predictor_comparison}) and the basis for evaluating the added predictive value of the extra hematological variables.

\section{Research Objectives and Scope}
The objectives of this research are the following: to review existing machine learning approaches for dengue prediction; to develop a benchmark prediction model using predictors common to both datasets; to develop an extended prediction model by incorporating additional clinical and hematological predictors; to compare the performance of Logistic Regression, Random Forest, and XGBoost; and to identify the most influential predictors for dengue infection.

These objectives are guided by three research questions: which machine learning model provides the best performance for dengue infection prediction; whether adding additional clinical and hematological predictors improves prediction performance and which clinical and hematological variables are the most important predictors of dengue infection.

The study is limited to binary classification (dengue-positive versus dengue-negative) of
hospital-recorded patients using routinely collected demographic and hematological variables.
It does not attempt to predict dengue severity, hospital length of stay or outcomes and it does
not use climate, seasonal or geographic information because neither dataset records admission
dates or patient locations. Three widely used classifiers (Logistic Regression, Random Forest
and XGBoost) are compared. Other model families such as support vector machines and neural
networks are outside the scope. Each dataset is modelled separately and no model is trained on
one year and tested on the other. The findings therefore describe predictive performance within
two Bangladeshi hospital-based datasets and should not be taken as evidence of performance in
other settings or populations.

\section{Methodology}

This study adopts a quantitative research approach using two publicly available Bangladeshi clinical and hematological dengue datasets during 2024 and 2025.

\subsection{Data Collection}

Two publicly available Bangladeshi clinical and hematological dengue datasets were used in this study, collected at different hospitals in different years. Their key characteristics are summarised below.\

\begin{itemize}[nosep]
    \item \textbf{Dataset 1 (2024):} Clinical and hematologic data collected between February and September 2024 ($n = 1{,}523$ raw records) from Jamalpur Medical College Hospital, Jamalpur, Bangladesh \citep{mendeley1}.
    \item \textbf{Dataset 2 (2025):} Clinical and hematological data collected between July and November 2025 ($n = 1{,}329$ raw records) across several medical institutions in Dhaka, Bangladesh, including Shaheed Suhrawardy Medical College and Hospital and Dhaka Medical College Hospital \citep{mendeley2}.
\end{itemize}

Both datasets contain one record for each patient, so they provide a snapshot of the patients rather than information over time. There is no admission date for individual patients and only the overall data collection period is known. Therefore, it was not possible to analyse changes over time or seasonal patterns.These datasets provide useful information for developing machine learning models for dengue infection prediction.

\subsection{Predictor Comparison}
\label{subsec:predictor_comparison}
Table~\ref{tab:predictors} summarises the correspondence between the two datasets' raw variables.

\begin{table}[h]
\centering
\caption{Predictor correspondence between Dataset 1 (2024) and Dataset 2 (2025)}
\label{tab:predictors}
\small
\begin{tabular}{L{2.6cm}L{2.6cm}L{2.6cm}L{5.5cm}}
\toprule
\textbf{Variable} & \textbf{Dataset 1 (2024)} & \textbf{Dataset 2 (2025)} & \textbf{Notes} \\
\midrule
Age & Age & Age & Common; D1 integer years; D2 has a 0.5 value (infant, in years) \\
Sex & Gender & Sex & Common; Male/Female, no missing \\
Red blood cell count & RBC & RBC & Common; Same scale (millions/µL) but D2 has extreme outliers (min 0.5, max 66.9) needing cleaning. \\
Haematocrit & HCT(\%) & HCT & Common; Both (\%),comparable range \\
Neutrophils \% & Neutrophils(\%) & Neut\% & Common \\
Lymphocytes \% & Lymphocytes(\%) & Lymph\% & Common \\
Platelet count & Total Platelet Count(/cumm) & PLT & Common; D1 raw cells/µL (56,000–299,803); D2 in ×10³/µL (0.02–897) — needs rescaling (D2 × 1000) before pooling. \\
WBC count & Total WBC count(/cumm) & WBC & Common; D1 raw cells/$\mu$L (3,500--14,900); D2 in $\times 10^3/\mu$L (0--48.1) --- same rescaling issue. \\
Haemoglobin & Hemoglobin(g/dl) & -- & 2024 only \\
Monocytes \% & Monocytes(\%) & -- & 2024 only \\
Eosinophils \% & Eosinophils(\%) & -- & 2024 only \\
MCV & MCV(fl) & -- & 2024 only \\
MCH & MCH(pg) & -- & 2024 only \\
MCHC & MCHC(g/dl) & -- & 2024 only \\
RDW-CV & RDW-CV(\%) & -- & 2024 only \\
MPV & MPV(fl) & -- & 2024 only \\
PDW & PDW(\%) & -- & 2024 only \\
PCT & PCT(\%) & -- & 2024 only \\
ALT & -- & ALT & 2025 only; 1,292/1,329 missing (97\%) not usable as a predictor. \\
AST & -- & AST & 2025 only; 99\% missing, unusable \\
\bottomrule
\end{tabular}
\end{table}

\begin{itemize}
    \item \textbf{Common predictors (8):} Age, Sex, RBC, HCT, Neutrophils\%, Lymphocytes\%, Platelet Count, WBC Count.

\item \textbf{2024-only predictors (10):} Hemoglobin, Monocytes\%, Eosinophils\%, MCV, MCH, MCHC, RDW-CV, MPV, PDW, PCT.

\item \textbf{2025-only variables (2, unusable):} ALT (97\% missing), AST (99\% missing) these 2 variables  excluded, since imputing over 97\% of a column would amount to fabricating data rather than modelling it.
\end{itemize}

The eight common predictors were retained for the benchmark model because they are the variables recorded consistently across both hospital sources and represent standard, low-cost demographic and hematological measures already established as informative in prior dengue studies (\citet{assaduzzaman2024}; \citet{nirob2025})  making them realistic candidates for a low-cost screening tool that does not depend on specialised serology. The ten additional predictors unique to the 2024 dataset were included in the extended model because each captures a distinct dimension of red-cell, white-cell or platelet morphology, for example, RDW-CV, MCV, MCH and MCHC describe red-cell size and haemoglobin content, while MPV and PDW describe platelet size and heterogeneity  that is plausibly altered by the plasma leakage and marrow suppression associated with dengue infection and that prior work has not evaluated in combination with the eight common predictors. The two variables unique to the 2025 dataset (ALT, AST) were excluded outright rather than retained as a third predictor set, since their extreme missingness (97–99\%) meant that any imputation would effectively be fabricating data rather than modelling an observed relationship.

\subsection{Data Preprocessing}

Before model training, both datasets were brought through a consistent sequence of cleaning steps. As a first step, column standardisation was carried out: each dataset's raw column names were mapped to a shared schema (for example, Gender/Sex → Sex, and Total Platelet Count(/cumm)/PLT → Platelet Count), so that corresponding variables could be identified consistently across the two sources. Columns that were not usable as predictors were then removed — ALT, AST and the serial-number index column (SN) were dropped from the 2025 dataset, since ALT and AST were both over 97\% missing. Twelve exact duplicate rows were removed from the 2024 dataset and remaining rows with missing values were dropped following the column removals above.

A unit-correction step was also required: the 2025 dataset's Platelet Count and WBC Count had been recorded in ×10³/µL and were rescaled (×1000) to match the raw cells/µL scale used in the 2024 dataset. Categorical variables were then encoded as binary,for example, Sex (male = 1, female = 0) and Result (positive = 1, negative = 0). Finally, outliers were addressed by winsorising (IQR-capping) all numeric predictors in both datasets, rather than removing them outright and without imposing hardcoded clinical thresholds; this handled extreme values such as 2025 Dataset's RBC outliers (0.5–66.9) and 2024 Dataset's implausible PCT values (up to 234) while preserving the rest of each affected patient's record.

After cleaning, the 2024 dataset contained 1,511 patients and the 2025 dataset contained 1,283 patients.

\subsection{Benchmark and Extended Model Formulation}
Consistent with the proposed methodology, a \textbf{benchmark model} was formulated using only the eight predictors common to both years:
\begin{equation}
\text{Result} \sim \text{Age} + \text{Sex} + \text{RBC} + \text{HCT} + \text{Neutrophils\%} + \text{Lymphocytes\%} + \text{Platelet Count} + \text{WBC Count}
\label{eq:benchmark}
\end{equation}

The same benchmark model was used separately for the 2024 and 2025 datasets.Analysing each year separately using the same predictors allows a clear and fair comparison between 2024 and 2025.

An \textbf{extended model} was formulated for the 2024 dataset only, incorporating additional clinical and hematological predictors available in one dataset.The performance of the benchmark and extended models were compared to evaluate the contribution of the additional predictors.
\begin{flalign}
\begin{aligned}
\text{Result} \sim \, & \text{Benchmark predictors} + \text{Hemoglobin} + \text{Monocytes \%} + \text{Eosinophils\%}
 + \text{MCV} + \text{MCH} + \\
\text{MCHC}& + \text{RDW-CV} + \text{MPV} + \text{PDW} + \text{PCT}
\end{aligned} &&
\label{eq:extended}
\end{flalign}

No extended model was developed for 2025, since its only additional variables ALT/AST are missing in over 97\% of cases, imputing them would mean fabricating data rather than modelling anything real. Dataset 2 is therefore only able to support the benchmark model.

Critically, the benchmark and extended datasets for 2024 were built from the \emph{same 1{,}511 patients} -- differing only in the number of predictor columns. This ensures that any performance difference between the benchmark and extended 2024 models reflects the contribution of the additional predictors alone, not a difference in sample composition.

\subsection{Machine Learning Model Development}
Three supervised algorithms were trained and compared in this study. \textbf{Logistic Regression} was used as a baseline linear model, providing an interpretable reference point against which the more complex algorithms could be evaluated. \textbf{Random Forest}, an ensemble of 300 decision trees was included for its ability to capture non-linear relationships and interactions between predictors without requiring them to be specified in advance. \textbf{XGBoost}, trained with 300 boosting rounds was also included as a gradient-boosted tree-based algorithm known for strong predictive performance on structured, tabular clinical data of this kind.
Prior to training, all numeric features were standardised to zero mean and unit variance, so that predictors measured on different scales would contribute comparably to model fitting rather than being dominated by variables with larger raw magnitudes. Each dataset was then split 80/20 into training and test sets using stratified sampling, to preserve the original class ratio in the test set. All experiments used a fixed random seed (42) to ensure reproducibility.

\subsection{Model Evaluation}
Model performance was assessed using Accuracy, Precision, Recall, F1-score, ROC-AUC and confusion matrices consistent with the metrics specified in the proposal. Confusion matrices summarise predictions into four counts: true positives (TP), true negatives (TN), false positives (FP) and false negatives (FN), from which the remaining metrics are derived.

Accuracy measures the overall proportion of correctly classified patients:

\begin{equation}
\text{Accuracy} = \frac{TP + TN}{TP + TN + FP + FN}
\label{eq:accuracy}
\end{equation}

Precision measures, among patients predicted positive for dengue, the proportion who were truly positive:

\begin{equation}
\text{Precision} = \frac{TP}{TP + FP}
\label{eq:precision}
\end{equation}

Recall (sensitivity) measures, among patients who were truly positive, the proportion correctly identified by the model:

\begin{equation}
\text{Recall} = \frac{TP}{TP + FN}
\label{eq:recall}
\end{equation}

F1-score is the harmonic mean of Precision and Recall, balancing the two into a single measure:

\begin{equation}
\text{F1-score} = 2 \times \frac{\text{Precision} \times \text{Recall}}{\text{Precision} + \text{Recall}}
\label{eq:f1}
\end{equation}

ROC-AUC summarises the trade-off between true-positive rate and false-positive rate across all classification thresholds, with a value of 1.0 indicating perfect discrimination and 0.5 indicating no better than chance.

Because both datasets in this study are imbalanced, with substantially more dengue-positive than dengue-negative patients (Section~\ref{subsec:addressing_class_imbalance}), Recall and Precision were also computed explicitly for the negative class, following the same definitions as Equations~\ref{eq:precision} and \ref{eq:recall} but with the positive and negative labels reversed:

\begin{equation}
\text{Recall (Negative)} = \frac{TN}{TN + FP}
\label{eq:recall_neg}
\end{equation}

\begin{equation}
\text{Precision (Negative)} = \frac{TN}{TN + FN}
\label{eq:precision_neg}
\end{equation}

These negative-class metrics are reported explicitly because standard Accuracy, Precision and Recall (Equations~\ref{eq:accuracy}–\ref{eq:recall}) can appear favourable even when a model performs poorly at identifying non-dengue patients — the specific failure mode under investigation in the class-imbalance analysis (Section~\ref{subsec:addressing_class_imbalance}).

\subsection{Addressing Class Imbalance}
\label{subsec:addressing_class_imbalance}
The two datasets had different class distributions. In 2024 68.5\% of patients tested positive for dengue and 31.5\% tested negative. In 2025 74.8\% tested positive and 25.2\% tested negative. Therefore, the 2025 dataset had about 6 percentage points more positive cases than 2024. Both datasets were imbalanced with more positive than negative cases. Without addressing this imbalance, the models tended to predict more patients as dengue-positive, leading to high sensitivity but lower specificity. (Section~\ref{sec:results}).\\
To address this imbalance, two resampling strategies were evaluated, applied exclusively to the training partition; the held-out 20\% test set was never resampled, so that model evaluation continued to reflect an honest, real-world class distribution. \textbf{SMOTE (Synthetic Minority Oversampling Technique)} generates synthetic examples of the minority (negative) class until class parity is achieved within the training data, whereas \textbf{Random undersampling} instead removes examples of the majority (positive) class until the two classes are equally represented.

Both approaches improved the models' capacity to identify negative cases relative to training on the unbalanced data directly. However, random undersampling produced a greater reduction in F1-score and ROC-AUC than SMOTE, attributable to the loss of genuine training observations from an already modest dataset. Consequently, SMOTE was adopted as the balancing method for all subsequent model results reported in this study.

\section{Results}
\label{sec:results}
Results are reported separately for the 2024 and 2025 datasets, given the substantial difference in their baseline predictive difficulty (discussed further in Section~\ref{sec:discussion}).

\subsection{2024 Results}

\subsubsection{Effect of Class Imbalance (2024 Benchmark)}
Table~\ref{tab:imbalance2024} compares no balancing, SMOTE and undersampling for the 2024 benchmark model.

\begin{table}[H]
\centering
\caption{2024 benchmark model: effect of class-imbalance handling}
\label{tab:imbalance2024}
\small
\begin{tabular}{lccccccccc}
\toprule
& \multicolumn{3}{c}{Recall (Negative)} & \multicolumn{3}{c}{F1-score} & \multicolumn{3}{c}{ROC-AUC} \\
\cmidrule(lr){2-4}\cmidrule(lr){5-7}\cmidrule(lr){8-10}
Model & None & SMOTE & Under & None & SMOTE & Under & None & SMOTE & Under \\
\midrule
Logistic Regression & 0.189 & 0.484 & 0.453 & 0.825 & 0.679 & 0.646 & 0.570 & 0.570 & 0.552 \\
Random Forest        & 0.274 & 0.347 & 0.463 & 0.834 & 0.816 & 0.711 & 0.635 & 0.640 & 0.610 \\
XGBoost               & 0.347 & 0.379 & 0.526 & 0.792 & 0.786 & 0.632 & 0.647 & 0.645 & 0.588 \\
\bottomrule
\end{tabular}
\end{table}

Without balancing, all three models struggled to correctly identify non-dengue patients, with negative-class recall ranging from only 18.9\% to 34.7\%. SMOTE greatly improved negative-class recall for Logistic Regression and Random Forest, while keeping their F1-score and ROC-AUC at a similar or slightly better level. Random undersampling gave the highest negative-class recall, but it reduced F1-score and ROC-AUC because it removed real training data from the relatively small 2024 dataset. Therefore, SMOTE was selected as the balancing method for the main 2024 results.

\subsubsection{2024 Benchmark Model (SMOTE-balanced)}
\begin{figure}[h]
\centering
\includegraphics[width=\textwidth]{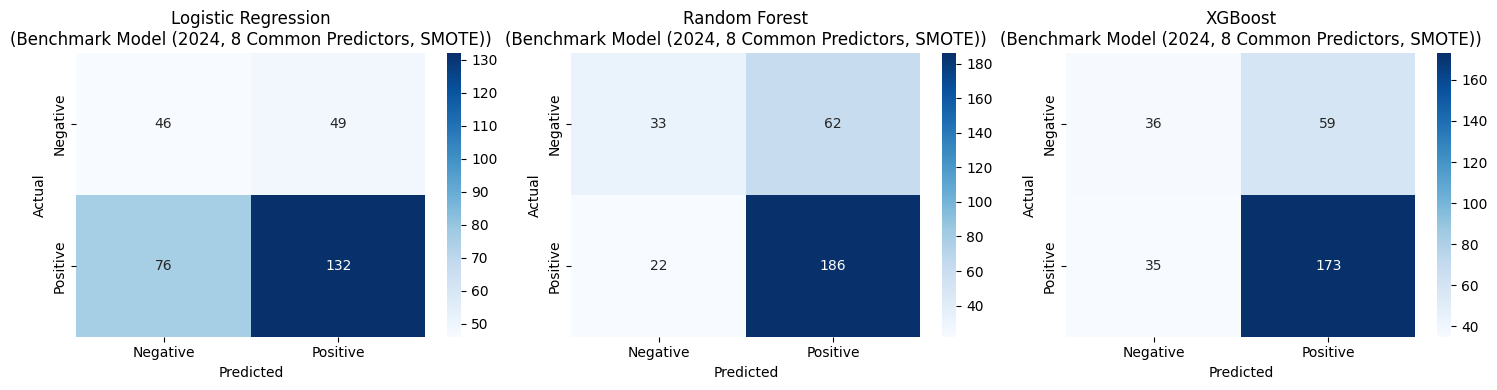}
\caption{Confusion matrices -- 2024 Benchmark Model (8 common predictors, SMOTE)}
\label{fig:cm2024}
\end{figure}

\begin{figure}[H]
\centering
\includegraphics[width=0.65\textwidth]{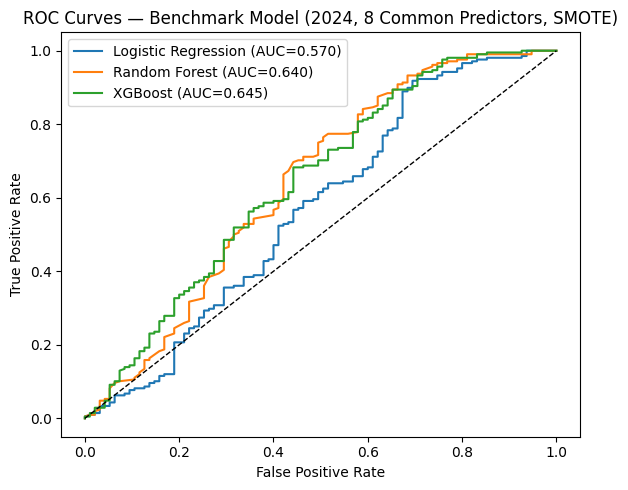}
\caption{ROC curves -- 2024 Benchmark Model (8 common predictors, SMOTE)}
\label{fig:roc2024}
\end{figure}

\begin{figure}[H]
  \centering
\includegraphics[width=0.75\linewidth]{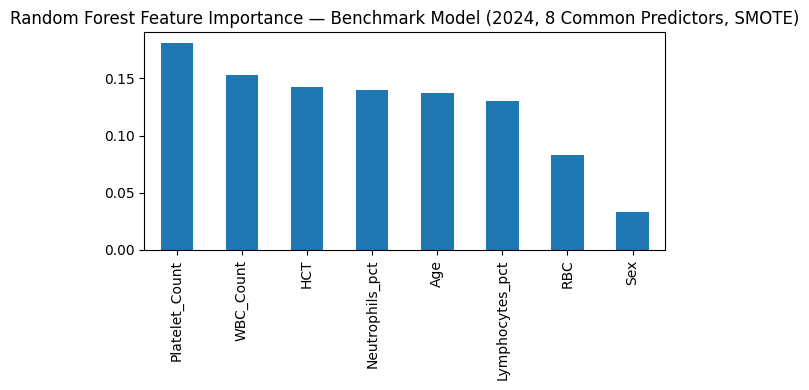}
  \caption{Feature importance for the 2024 benchmark model.}
  \label{fig:fi_benchmark2024}
\end{figure}

Figure~\ref{fig:cm2024} shows the confusion matrices and Figure~\ref{fig:roc2024} the ROC curves for the 2024 benchmark model (8 predictors, SMOTE-balanced) and Figure~\ref{fig:fi_benchmark2024} the Random Forest
feature-importance ranking for the benchmark model. Table~\ref{tab:bench2024} reports the full metrics.

\begin{table}[h]
\centering
\caption{2024 Benchmark Model performance (8 common predictors, SMOTE-balanced)}
\label{tab:bench2024}
\small
\begin{tabular}{lccccccc}
\toprule
Model & Accuracy & Precision & Recall & F1-score & ROC-AUC & Recall (Neg) & Precision (Neg) \\
\midrule
Logistic Regression & 0.587 & 0.729 & 0.635 & 0.679 & 0.570 & 0.484 & 0.377 \\
Random Forest        & 0.723 & 0.750 & 0.894 & 0.816 & 0.640 & 0.347 & 0.600 \\
XGBoost               & 0.690 & 0.746 & 0.832 & 0.786 & 0.645 & 0.379 & 0.507 \\
\bottomrule
\end{tabular}
\end{table}

Random Forest achieves the highest F1-score (0.816) and Accuracy (0.723); XGBoost achieves the highest ROC-AUC (0.645), indicating marginally better discrimination across all classification thresholds despite a lower F1-score at the default 0.5 threshold.

\subsubsection{2024 Extended Model (18 predictors, SMOTE-balanced)}
\begin{figure}[h]
\centering
\includegraphics[width=\textwidth]{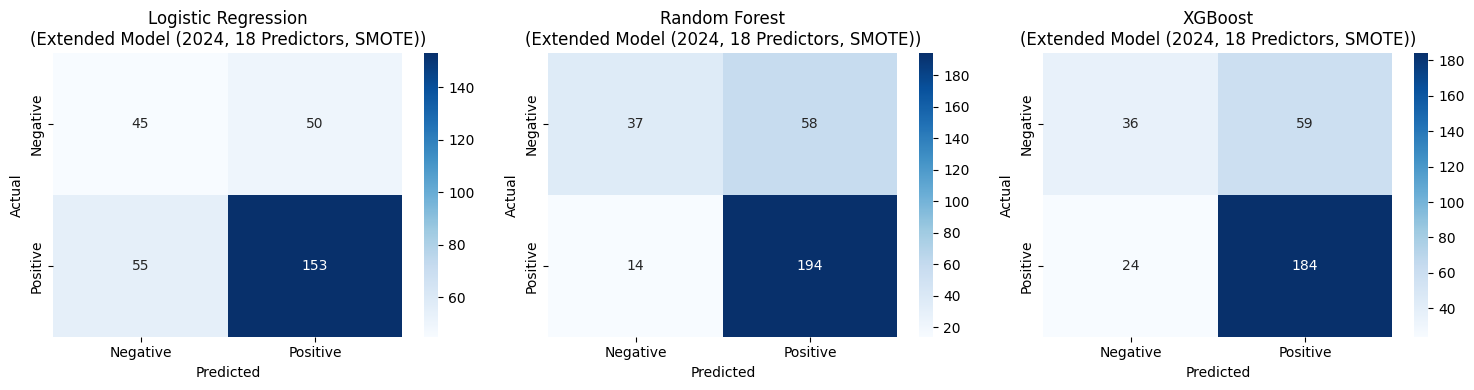}
\caption{Confusion matrices -- 2024 Extended Model (18 predictors, SMOTE)}
\label{fig:cmext}
\end{figure}

\begin{figure}[h]
\centering
\includegraphics[width=0.65\textwidth]{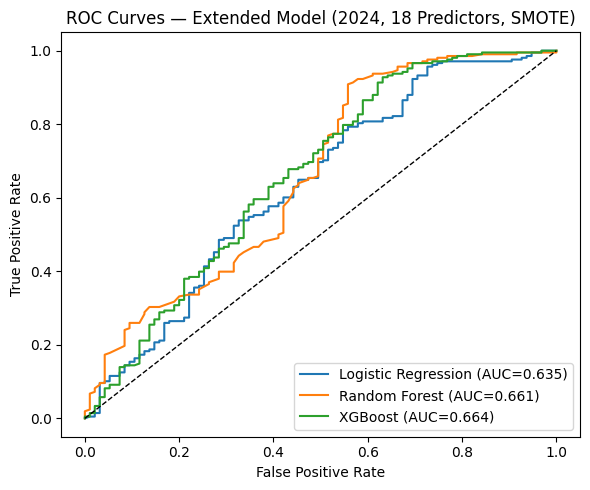}
\caption{ROC curves -- 2024 Extended Model (18 predictors, SMOTE)}
\label{fig:rocext}
\end{figure}

\begin{figure}[H]
\centering
\includegraphics[width=0.7\textwidth]{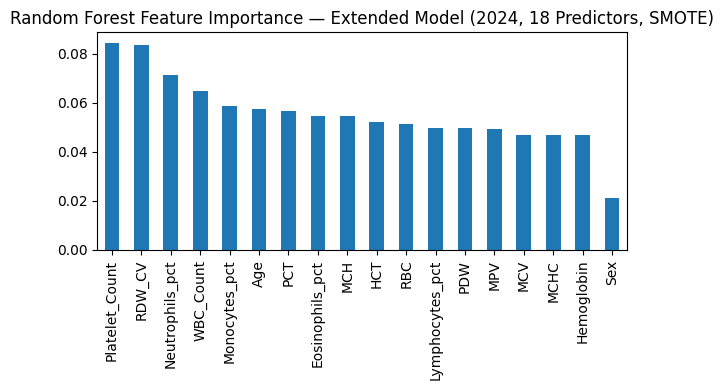}
\caption{Random Forest feature importance -- 2024 Extended Model (18 predictors, SMOTE)}
\label{fig:fiext}
\end{figure}

Figure~\ref{fig:cmext} shows the confusion matrices, Figure~\ref{fig:rocext} the ROC curves and Figure~\ref{fig:fiext} the Random Forest feature-importance ranking for the extended model. Table~\ref{tab:ext2024} reports the full metrics.

\begin{table}[H]
\centering
\caption{2024 Extended Model performance (18 predictors, SMOTE-balanced)}
\label{tab:ext2024}
\small
\begin{tabular}{lccccccc}
\toprule
Model & Accuracy & Precision & Recall & F1-score & ROC-AUC & Recall (Neg) & Precision (Neg) \\
\midrule
Logistic Regression & 0.653 & 0.754 & 0.736 & 0.745 & 0.635 & 0.474 & 0.450 \\
Random Forest        & 0.762 & 0.770 & 0.933 & 0.843 & 0.661 & 0.389 & 0.725 \\
XGBoost               & 0.726 & 0.757 & 0.885 & 0.816 & 0.664 & 0.379 & 0.600 \\
\bottomrule
\end{tabular}
\end{table}

\subsubsection{Within-2024 Comparison: Benchmark vs Extended}
Table~\ref{tab:within2024} isolates the effect of the ten additional predictors, using identical patients and an identical train/test split, so that the comparison reflects feature contribution alone.

\begin{table}[h]
\centering
\caption{2024: Benchmark (8 predictors) vs Extended (18 predictors), identical rows/split}
\label{tab:within2024}
\small
\begin{tabular}{lcccccc}
\toprule
Model & \multicolumn{2}{c}{F1-score} & \multicolumn{1}{c}{$\Delta$} & \multicolumn{2}{c}{ROC-AUC} & \multicolumn{1}{c}{$\Delta$} \\
& Benchmark & Extended & F1 & Benchmark & Extended & AUC \\
\midrule
Logistic Regression & 0.679 & 0.745 & +0.066 & 0.570 & 0.635 & +0.065 \\
Random Forest        & 0.816 & 0.843 & +0.027 & 0.640 & 0.661 & +0.021 \\
XGBoost               & 0.786 & 0.816 & +0.030 & 0.645 & 0.664 & +0.019 \\
\bottomrule
\end{tabular}
\end{table}

Every algorithm improves on both F1-score and ROC-AUC when the ten additional hematological predictors are included, using an identical sample and split. Because the sample and train/test split were the same, the improvement supports the contribution of the additional predictors to dengue classification.

\subsection{2025 Results}

\subsubsection{Effect of Class Imbalance (2025 Benchmark)}
Table~\ref{tab:imbalance2025} compares no balancing, SMOTE and undersampling for the 2025 benchmark model.

\begin{table}[h]
\centering
\caption{2025 benchmark model: effect of class-imbalance handling}
\label{tab:imbalance2025}
\small
\begin{tabular}{lccccccccc}
\toprule
& \multicolumn{3}{c}{Recall (Negative)} & \multicolumn{3}{c}{F1-score} & \multicolumn{3}{c}{ROC-AUC} \\
\cmidrule(lr){2-4}\cmidrule(lr){5-7}\cmidrule(lr){8-10}
Model & None & SMOTE & Under & None & SMOTE & Under & None & SMOTE & Under \\
\midrule
Logistic Regression & 0.446 & 0.785 & 0.769 & 0.891 & 0.874 & 0.887 & 0.844 & 0.848 & 0.857 \\
Random Forest        & 0.631 & 0.723 & 0.785 & 0.920 & 0.915 & 0.895 & 0.895 & 0.904 & 0.905 \\
XGBoost               & 0.692 & 0.677 & 0.769 & 0.929 & 0.919 & 0.860 & 0.903 & 0.902 & 0.892 \\
\bottomrule
\end{tabular}
\end{table}

2025's class imbalance had a smaller effect than 2024's. Even without any balancing, negative-class recall was already much higher for 2025 (44.6\%–69.2\%) than it was for 2024. Both SMOTE and undersampling improved this further and this time with less cost to F1-score and ROC-AUC than seen in 2024. SMOTE was kept for consistency with the 2024 results.

\subsubsection{2025 Benchmark Model (SMOTE-balanced)}
\begin{figure}[h]
\centering
\includegraphics[width=\textwidth]{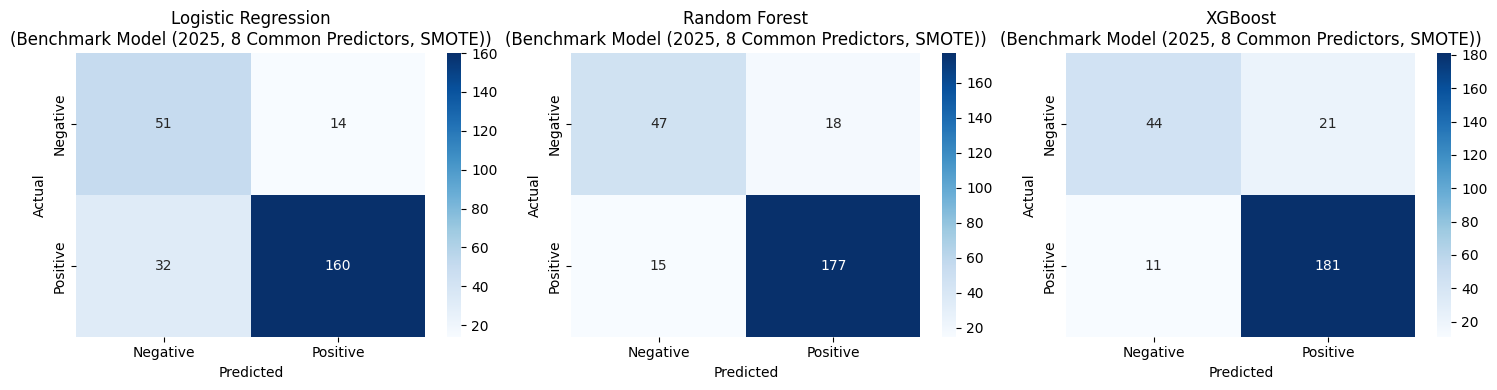}
\caption{Confusion matrices -- 2025 Benchmark Model (8 common predictors, SMOTE)}
\label{fig:cm2025}
\end{figure}

\begin{figure}[h]
\centering
\includegraphics[width=0.65\textwidth]{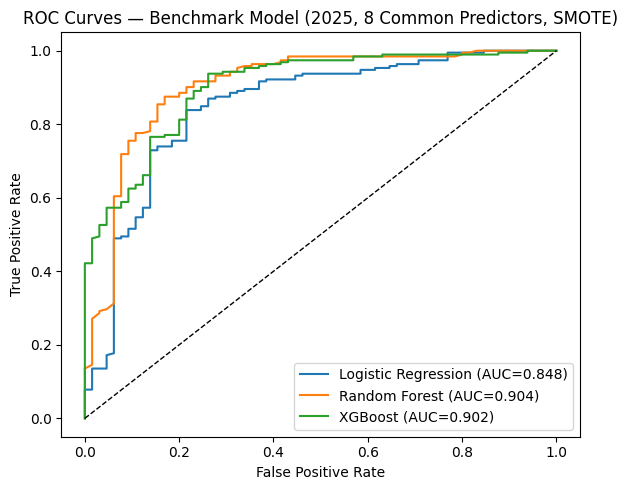}
\caption{ROC curves -- 2025 Benchmark Model (8 common predictors, SMOTE)}
\label{fig:roc2025}
\end{figure}

Figure~\ref{fig:cm2025} and Figure~\ref{fig:roc2025} show the confusion matrices and ROC curves; Table~\ref{tab:bench2025} reports the full metrics.

\begin{table}[h]
\centering
\caption{2025 Benchmark Model performance (8 common predictors, SMOTE-balanced)}
\label{tab:bench2025}
\small
\begin{tabular}{lccccccc}
\toprule
Model & Accuracy & Precision & Recall & F1-score & ROC-AUC & Recall (Neg) & Precision (Neg) \\
\midrule
Logistic Regression & 0.821 & 0.920 & 0.833 & 0.874 & 0.848 & 0.785 & 0.614 \\
Random Forest        & 0.872 & 0.908 & 0.922 & 0.915 & 0.904 & 0.723 & 0.758 \\
XGBoost               & 0.875 & 0.896 & 0.943 & 0.919 & 0.902 & 0.677 & 0.800 \\
\bottomrule
\end{tabular}
\end{table}

XGBoost achieves the highest F1-score (0.919) and Accuracy (0.875); Random Forest achieves the highest ROC-AUC (0.904) and the best negative-class recall (0.723) among the tree-based models. All three algorithms substantially outperform their 2024 counterparts on every metric.

\subsection{Cross-Year Comparison and Predictor Importance}

\subsubsection{2024 vs 2025 (Identical 8 Predictors)}
Table~\ref{tab:yearcompare} compares the benchmark models across years.

\begin{table}[h]
\centering
\caption{Benchmark model comparison: 2024 vs 2025 (identical 8 predictors)}
\label{tab:yearcompare}
\small
\begin{tabular}{lcccc}
\toprule
Model & 2024 F1 & 2025 F1 & 2024 ROC-AUC & 2025 ROC-AUC \\
\midrule
Logistic Regression & 0.679 & 0.874 & 0.570 & 0.848 \\
Random Forest        & 0.816 & 0.915 & 0.640 & 0.904 \\
XGBoost               & 0.786 & 0.919 & 0.645 & 0.902 \\
\bottomrule
\end{tabular}
\end{table}

Despite using an identical predictor set, ROC-AUC is 0.256--0.274 higher for 2025 than for 2024 across all three algorithms -- discussed in detail in Section~\ref{sec:discussion}.

\subsubsection{Most Influential Extended Predictors}
Table~\ref{tab:importance} ranks the ten 2024-only predictors by importance across all three algorithms (Random Forest and XGBoost importances; absolute Logistic Regression coefficient magnitude), together with the mean rank across models.

\begin{table}[h]
\centering
\caption{Extended predictors ranked by mean importance rank across three models}
\label{tab:importance}
\small
\begin{tabular}{lcccc}
\toprule
Predictor & Random Forest & XGBoost & LR $|$coef$|$ & Mean rank \\
\midrule
Monocytes\%      & 0.0587 & 0.1043 & 0.3012 & 1.67 \\
Eosinophils\%    & 0.0546 & 0.1283 & 0.2005 & 2.33 \\
RDW-CV           & 0.0835 & 0.0672 & 0.1585 & 3.00 \\
MCH              & 0.0545 & 0.0419 & 0.1853 & 4.00 \\
PCT              & 0.0566 & 0.0386 & 0.1516 & 4.67 \\
MPV              & 0.0494 & 0.0317 & 0.1345 & 7.00 \\
PDW              & 0.0496 & 0.0350 & 0.0157 & 7.33 \\
MCHC             & 0.0468 & 0.0281 & 0.1821 & 7.33 \\
MCV              & 0.0468 & 0.0262 & 0.0470 & 8.67 \\
Hemoglobin       & 0.0467 & 0.0317 & 0.0388 & 9.00 \\
\bottomrule
\end{tabular}
\end{table}

Monocytes\% and Eosinophils\% had the highest average ranks, but the models differed considerably in their individual rankings: the three models disagree substantially on how important they are. Eosinophils\%, for example, ranks 4th by Random Forest but 1st by XGBoost.
\textbf{RDW-CV} is different: it lands 1st by Random Forest, 3rd by XGBoost, and 5th by Logistic Regression — never dropping out of the upper half of the ranking for any model, even though its individual rank shifts more than the average-importance leaders do. In the full 18-predictor Random Forest ranking (Figure~\ref{fig:fiext}), it comes second overall just behind Platelet Count. This combination — a solid mean rank (3.00) and a presence among the stronger predictors in every model — makes RDW-CV the most defensible single finding on which added predictor matters, though it is not the predictor with the least rank variation across models (that distinction belongs to MPV, whose importance is consistently low rather than consistently high). It also fits what would be clinically expected: dengue can cause plasma leakage and changes in blood concentration, which may increase variation in red blood cell size. RDW-CV measures this variation, making it a biologically plausible predictor for dengue.

\section{Discussion}
\label{sec:discussion}

No single model performed best across all datasets and evaluation measures.Random Forest gave the most consistent results, achieving the best or nearly the best F1-score in both years after balancing. Its F1-score was 0.816 for the 2024 benchmark model, 0.843 for the 2024 extended model, and 0.915 for the 2025 benchmark model, just below XGBoost (0.919). XGBoost achieved the highest ROC-AUC in most experiments, showing slightly better ability to distinguish between dengue-positive and dengue-negative patients. Logistic Regression was the weakest model overall, possibly because the relationship between blood-related predictors and dengue status is not simply linear. Overall, Random Forest was considered the most suitable model because it provided consistently strong F1-scores, competitive ROC-AUC, and stable performance after addressing class imbalance correction (Tables~\ref{tab:imbalance2024}--\ref{tab:imbalance2025}) relative to Logistic Regression.

Within the 2024 dataset, using the same patients and the same train/test split (Table~\ref{tab:within2024}), every model improved on both F1-score and ROC-AUC once the ten additional hematological predictors were added. Since the sample size and composition stayed exactly the same, the only thing that changed was the number of predictors, this makes it a fair, controlled comparison. The improvement was largest for Logistic Regression (ROC-AUC +0.065) and smaller for the tree-based models (Random Forest +0.021; XGBoost +0.019). As Logistic Regression is a linear model, it benefits more directly from new, individually useful features, while Random Forest and XGBoost can already pick up some non-linear patterns among the original eight predictors even without the extra ones.

For the eight shared predictors, Platelet Count is clearly the most important in the 2024 benchmark model (Figure~\ref{fig:fi_benchmark2024}), followed by WBC Count, HCT, Neutrophils\%, and Age at similar levels, with RBC and Sex contributing least. This fits what is clinically expected, since platelet and white blood cell counts are well-known markers of dengue severity.\\
Platelet Count and RDW-CV were consistently the two most important predictors in the 2024 extended Random Forest model (Figure~\ref{fig:fiext}), together with WBC Count and Neutrophils\%. Among the ten predictors unique to the extended model,RDW-CV was the only one to rank consistently highly across all three algorithms (Table~\ref{tab:importance}), making it the most defensible answer to which clinical and hematological variables are the most important predictors of dengue infection. Monocytes\% and Eosinophils\% also scored highly by mean rank but showed model-dependent inconsistency and are reported as secondary, less certain findings.

Taken together, Platelet Count is the single most important predictor across both the benchmark and extended models, while RDW-CV is the most valuable addition once the extended predictor set is introduced.

\subsection{Comparison with Previous Research}
The closest prior work is \citet{tuhin2025}, whose stacking ensemble (DengueStackX-19) was built on the same 1,523-patient source as the 2024 dataset used here and reported 93.65\% accuracy and 89.63\% F1-score under 10-fold cross-validation. The best 2024 results in this study are considerably lower (Random Forest accuracy 0.723, F1-score 0.816, ROC-AUC 0.640 for the benchmark model; 0.762, 0.843 and 0.661 for the extended model). Several differences may contribute: their model used stacking, SMOTEENN and a richer set of engineered features and was evaluated by cross-validation whereas this study used plain SMOTE, three single models and a single held-out test set. The figures are therefore indicative rather than strictly comparable and the gap should not be read as showing that either approach is superior. \citet{haque2025} reported 99.49\% testing accuracy with a genetic-algorithm and GAN-based framework; the 2025 benchmark results here (Random Forest accuracy 0.872, F1-score 0.915, ROC-AUC 0.904) are lower, but they were obtained with only eight common predictors and a test set containing no synthetic data. Very high accuracy figures may partly reflect fitting to a single localised dataset which is consistent with the sharp 2024--2025 contrast observed here.

The finding that Platelet Count is the most important predictor in the 2024 benchmark model agrees with \citet{dasgupta2025}, who also found platelet count to be the most influential predictor and is consistent with the clinical description of hematological changes in dengue by \citet{hossain2025}. Most earlier studies rely on a single dataset \citep{tuhin2025,haque2025,qaiser2024,bohm2024}; this study shows that the same predictors can behave very differently across datasets, since Platelet Count had an individual AUC of 0.528 in 2024 but 0.870 in 2025. The benchmark-versus-extended comparison and the identification of RDW-CV as the most consistent added predictor address a gap that the literature review did not find covered elsewhere, although the RDW-CV finding comes from a single dataset and should be confirmed independently.

\subsection{The 2024--2025 Performance Gap}
A substantial and consistent gap was observed between the 2024 and 2025 benchmark models using \emph{identical} predictors (Table~\ref{tab:yearcompare}): ROC-AUC differed by 0.256--0.274 across all three algorithms, with 2025 performing considerably better. The possible reasons for this difference were then examined.

Class imbalance does not appear to be the primary driver. If imbalance alone explained the gap, correcting it via SMOTE should have narrowed the difference substantially. In practice, SMOTE changed 2024  Random Forest ROC-AUC by only $+0.005$ (0.635 $\rightarrow$ 0.640; Table~\ref{tab:imbalance2024}), a negligible shift relative to the $\sim$0.26 year-to-year gap. The same pattern holds in 2025: Random Forest’s ROC-AUC moves by only +0.009 under SMOTE (0.895 → 0.904; Table~\ref{tab:imbalance2025}).This suggests the two datasets differ in some more fundamental respect than class proportion alone.

\begin{table}[h]
\centering
\caption{Individual predictor discriminative power (univariate ROC-AUC against Result), by year}
\label{tab:individualauc}
\small
\begin{tabular}{lcc}
\toprule
Predictor & 2024 individual AUC & 2025 individual AUC \\
\midrule
Age              & 0.553 & 0.564 \\
RBC              & 0.583 & 0.531 \\
HCT              & 0.544 & 0.546 \\
Neutrophils\%    & 0.571 & 0.555 \\
Lymphocytes\%    & 0.580 & 0.560 \\
Platelet Count   & \textbf{0.528} & \textbf{0.870} \\
WBC Count        & 0.558 & 0.668 \\
\bottomrule
\end{tabular}
\end{table}

A supplementary diagnostic was conducted to investigate this further: the discriminative power of each individual predictor (its own univariate ROC-AUC against the outcome, computed independently in each year) was compared across the two datasets (Table~\ref{tab:individualauc}). Platelet Count consistently the single most important predictor in both years' Random Forest models, showed a striking difference: an individual AUC of only 0.528 in 2024 (barely distinguishable from chance) versus 0.870 in 2025. WBC Count showed a smaller but notable gap (0.558 versus 0.668), while the remaining common predictors (Age, RBC, HCT, Neutrophils\%, Lymphocytes\%) showed comparatively similar, modest discriminative power in both years (individual AUCs in the approximately 0.53–0.58 range for both datasets). Since the overall models rely heavily on Platelet Count, its near total loss of individual discriminative signal in the 2024 data plausibly may explain part of the  observed year to year performance gap.

\begin{table}[h]
\centering
\caption{Measurement resolution: number of distinct values per predictor, by year}
\label{tab:granularity}
\small
\begin{tabular}{lccl}
\toprule
Predictor & 2024 distinct values ($n=1{,}511$) & 2025 distinct values ($n=1{,}283$) & Coarser in \\
\midrule
Age              & 67   & 59  & 2025 \\
RBC              & 4    & 480 & 2024 \\
HCT              & 829  & 392 & 2025 \\
Neutrophils\%    & 23   & 183 & 2024 \\
Lymphocytes\%    & 23   & 172 & 2024 \\
Platelet Count   & 1{,}157 & 332 & 2025 \\
WBC Count        & 893  & 694 & 2025 \\
\bottomrule
\end{tabular}
\end{table}

An additional,Table~\ref{tab:granularity} compares the number of distinct values each common predictor takes per year. The pattern is mixed: RBC, Neutrophils\%, and Lymphocytes\% are recorded far more coarsely in 2024 (e.g., RBC takes only 4 distinct values in 2024 versus 480 in 2025), consistent with a coarser recording convention at that source. However, Age, HCT, Platelet Count and WBC Count show the opposite pattern, with more distinct values in 2024. Notably, Platelet Count -- the predictor most responsible for the year-to-year performance gap -- is recorded at higher resolution in 2024 (1,157 values) than in 2025 (332), so its loss of discriminative power cannot be explained by coarse rounding. Measurement resolution therefore offers a partial explanation for some predictors, but not for the study's central finding, which remains an open question for future data audits. This is noted as a data-quality limitation rather than a modelling deficiency and is a reasonable direction for future investigation rather than a claim that can be fully resolved with the data available.

\section{Limitations and Future Work}
Several limitations of this study should be acknowledged. Neither dataset recorded the individual patient's admission date, precluding any seasonal or climate-linked analysis; this represents a deliberate scope decision rather than an oversight. The 2024–2025 performance gap, while investigated in detail, cannot be fully attributed to a single cause with the data available: measurement resolution differences plausibly account for part of the gap in some predictors (RBC, Neutrophils\%, Lymphocytes\%), but not for Platelet Count, the predictor most responsible for the gap, the cause of which remains unresolved. Furthermore, no extended model could be developed for the 2025 dataset, as its only unique variables (ALT, AST) were missing in over 97\% of records. Finally, SMOTE generates synthetic training examples and although the test set was never resampled, some sensitivity of the reported results to the choice of balancing method was observed (Tables~\ref{tab:imbalance2024}–\ref{tab:imbalance2025}); this should be borne in mind when interpreting the metrics presented in this study.

Complementary approaches could address these limitations: external validation (training on one dataset and testing on the other), repeated stratified cross-validation with confidence intervals, probability calibration and decision-curve analysis and a formal data audit of the two sources.Table~\ref{tab:futurework} links each uncertainty to the question it raises and the study that would address it.

\begin{table}[H]
\centering
\caption{Open uncertainties and proposed follow-up designs}
\label{tab:futurework}
\small
\begin{tabular}{L{3.6cm}L{4.4cm}L{6.2cm}}
\toprule
Uncertainty & Question raised & Proposed design \\
\midrule
Cause of the 2024--2025 gap & Does it reflect data quality, case mix or recording practice? & Data audit with the source hospitals; compare patient characteristics and laboratory methods \\
Transferability & Does a model trained on one dataset work on the other? & Cross-dataset training and testing on the eight common predictors \\
Stability of estimates & How much do results depend on one split? & Repeated stratified cross-validation with confidence intervals \\
Role of RDW-CV & Does it add value beyond 2024? & Independent validation in a dataset recording RDW-CV \\
\bottomrule
\end{tabular}
\end{table}

Several opportunities follow from this work. Models could be validated prospectively in a hospital setting to establish how they perform on newly presenting patients. Datasets that record admission dates and locations would allow seasonal and geographic information to be combined with laboratory values. The same predictors could be used to predict dengue severity, so that high-risk patients are identified early. Finally, a short survey or interviews with clinicians could show what level of accuracy and explainability would make such a tool acceptable in practice.

\section{Conclusion}
This study developed and compared benchmark and extended machine learning models for dengue infection prediction using two independent Bangladeshi clinical and hematological datasets.Random Forest showed the most consistently strong performance across the analyses, while the extended predictor set improved prediction performance in the controlled within-year comparisons. RDW-CV was also identified as the most consistently influential additional predictor. The substantial performance difference between the 2024 and 2025 datasets was more closely associated with reduced discriminative power of individual predictors, particularly Platelet Count, than with class imbalance. These findings support the use of machine learning to assist early, non-invasive dengue diagnosis using routinely collected hematological data, while  highlighting the importance of data quality, predictor stability, and measurement consistency when models are developed or evaluated across different clinical datasets.

\clearpage
\addcontentsline{toc}{section}{References}
\nocite{mendeley1,mendeley2}
\bibliographystyle{unsrtnat}
\bibliography{references}

\newpage
\appendix
\section*{Appendix A: Class Balance Summary}
\addcontentsline{toc}{section}{Appendix A: Class Balance Summary}
\begin{table}[h]
\centering
\caption{Dataset sizes and class balance after preprocessing}
\begin{tabular}{lccc}
\toprule
Dataset & $n$ & Positive (\%) & Negative (\%) \\
\midrule
2024 Benchmark (8 predictors) & 1{,}511 & 68.5 & 31.5 \\
2024 Extended (18 predictors) & 1{,}511 & 68.5 & 31.5 \\
2025 Benchmark (8 predictors) & 1{,}283 & 74.8 & 25.2 \\
\bottomrule
\end{tabular}
\end{table}

\section*{Appendix B: Software and Tools}
\addcontentsline{toc}{section}{Appendix B: Software and Tools}
\begin{table}[H]
\centering
\renewcommand{\arraystretch}{1.3}
\caption{Software and tools used for data processing, modelling, and evaluation}
\label{tab:software_tools}
\begin{tabular}{@{}ll@{}}
\toprule
\textbf{Task} & \textbf{Library} \\
\midrule
Data handling and cleaning       & pandas, numpy \\
Data visualisation               & matplotlib, seaborn \\
Machine learning models          & scikit-learn \\
XGBoost model                    & xgboost \\
Class imbalance handling         & imbalanced-learn (SMOTE, RandomUnderSampler) \\
Model evaluation                 & sklearn.metrics \\
\midrule
\multicolumn{2}{@{}p{0.9\textwidth}@{}}{\footnotesize \textit{Programming language and environment: Python 3, Google Colab / Jupyter Notebook.}} \\
\bottomrule
\end{tabular}
\end{table}

\clearpage
\section*{Appendix C: Google Colab Code}
\label{sec:appendix_c}

\includegraphics[page=1, trim=1cm 1cm 1cm 1cm, clip, width=\textwidth]{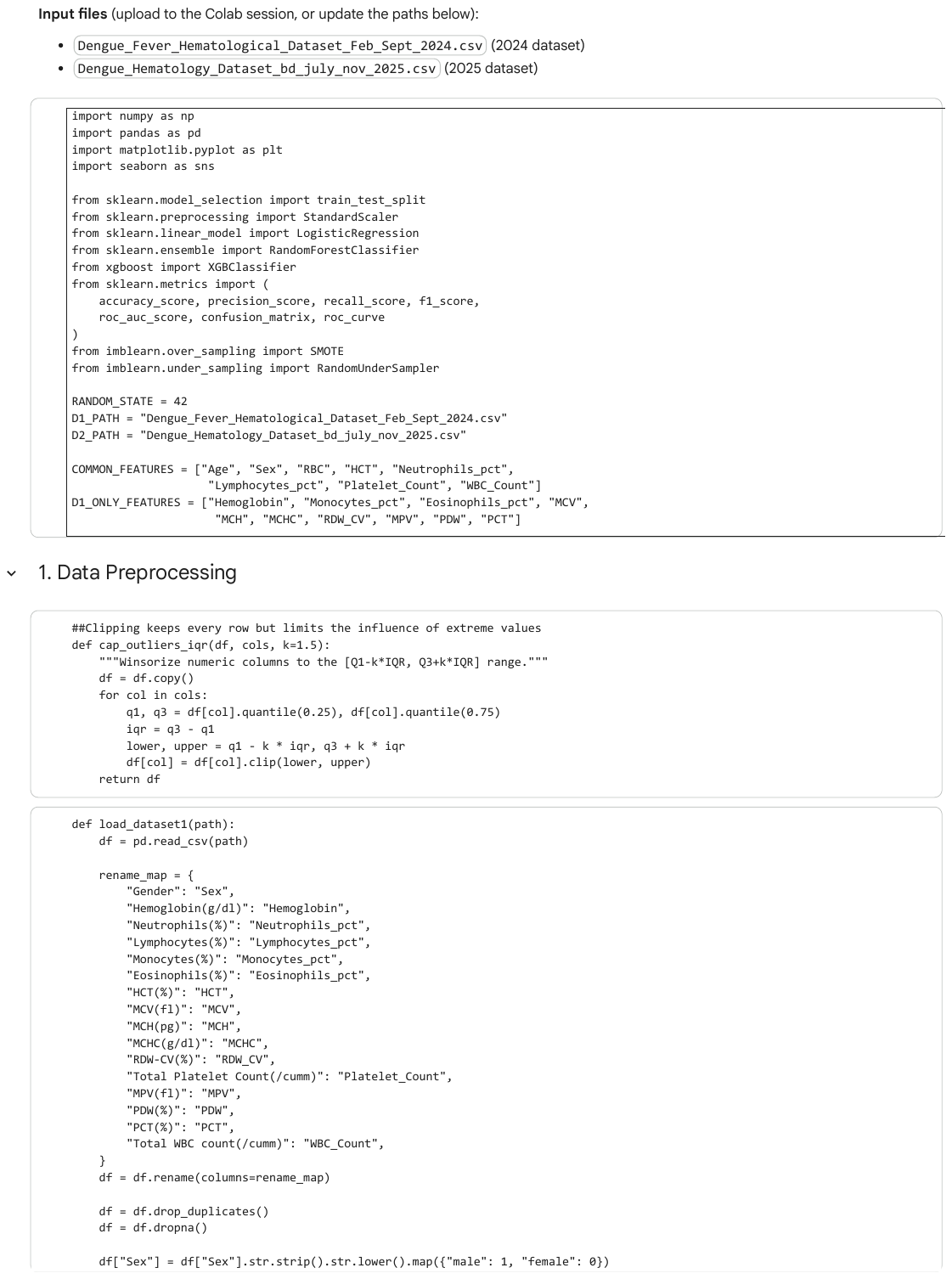}

\includegraphics[page=2, trim=1cm 1cm 1cm 1cm, clip, width=\textwidth]{ColabCode.pdf}

\includegraphics[page=3, trim=1cm 1cm 1cm 1cm, clip, width=\textwidth]{ColabCode.pdf}

\includegraphics[page=4, trim=1cm 1cm 1cm 1cm, clip, width=\textwidth]{ColabCode.pdf}

\includegraphics[page=5, trim=1cm 1cm 1cm 1cm, clip, width=\textwidth]{ColabCode.pdf}

\includegraphics[page=6, trim=1cm 1cm 1cm 1cm, clip, width=\textwidth]{ColabCode.pdf}

\includegraphics[page=7, trim=1cm 1cm 1cm 1cm, clip, width=\textwidth]{ColabCode.pdf}

\end{document}